\documentclass[10pt,twocolumn,letterpapeser]{article}

\usepackage{iccv}
\usepackage{times}
\usepackage{epsfig}
\usepackage{graphicx}
\usepackage{amsmath}
\usepackage{amssymb}
\usepackage{subfigure}
\newcommand{\tabincell}[2]{\begin{tabular}{@{}#1@{}}#2\end{tabular}}
\usepackage[normalem]{ulem}
\usepackage{multirow}
\usepackage{makecell}

\usepackage[accsupp]{axessibility}
\renewcommand{\thefootnote}{}
\usepackage[hang,flushmargin]{footmisc}
\usepackage{pdfpages}

\usepackage[pagebackref=true,breaklinks=true,letterpaper=true,colorlinks,bookmarks=false]{hyperref}

\iccvfinalcopy 

\def\iccvPaperID{1181} 

\ificcvfinal\fi

\begin{document}

\title{Seminar Learning for Click-Level Weakly Supervised Semantic Segmentation}

\author{Hongjun Chen$^{1}$, Jinbao Wang$^{1}$, Hong Cai Chen$^{1}$, Xiantong Zhen$^{2}$, \\Feng Zheng$^{1*}$, Rongrong Ji$^{3}$, Ling Shao$^{4}$ \\
$^1$ Southern University of Science and Technology \ $^2$ University of Amsterdam \\ $^3$ Xiamen University\  $^4$ Inception Institute of Artificial Intelligence 
}

\maketitle
\ificcvfinal\thispagestyle{empty}\fi

\let\thefootnote\relax\footnotetext{$^*$Corresponding author: Feng Zheng (Email: f.zheng@ieee.org). This work is supported by the National Natural Science Foundation of China under Grant No. 61972188.}

\begin{abstract}
Annotation burden has become one of the biggest barriers to semantic segmentation. Approaches based on click-level annotations have therefore attracted increasing attention due to their superior trade-off between supervision and annotation cost. In this paper, we propose \textit{seminar learning}, a new learning paradigm for semantic segmentation with click-level supervision. The fundamental rationale of seminar learning is to leverage the knowledge from different networks to compensate for insufficient information provided in click-level annotations. 
Mimicking a seminar, our seminar learning involves a teacher-student and a student-student module, where a student can learn from both skillful teachers and other students.
The teacher-student module uses a teacher network based on the exponential moving average to guide the training of the student network. In the student-student module, heterogeneous pseudo-labels are proposed to bridge the transfer of knowledge among students to enhance each other's performance.
Experimental results demonstrate the effectiveness of seminar learning, which achieves the new state-of-the-art performance of 72.51\% (mIOU), surpassing previous methods by a large margin of up to 16.88\% on the Pascal VOC 2012 dataset.
\end{abstract}

\section{Introduction}
Semantic segmentation is a fundamental task, where each pixel of an image is labeled into a predefined set of classes.  In the field of computer vision, it has made great progress in many applications, such as automatic driving, scene understanding, and medical diagnosis \cite{minaee2020image} \cite{2019A}. Recently, deep convolutional neural networks (CNNs) have achieved remarkable success in a variety of semantic segmentation tasks \cite{chen2017deeplab, minaee2020image}. However, they require large amounts of pixel-level annotations for training. The acquisition process of pixel-level annotations is extremely time-consuming and labor-intensive.

\begin{figure}[t]
	\begin{center}
		\includegraphics[width=0.9\linewidth]{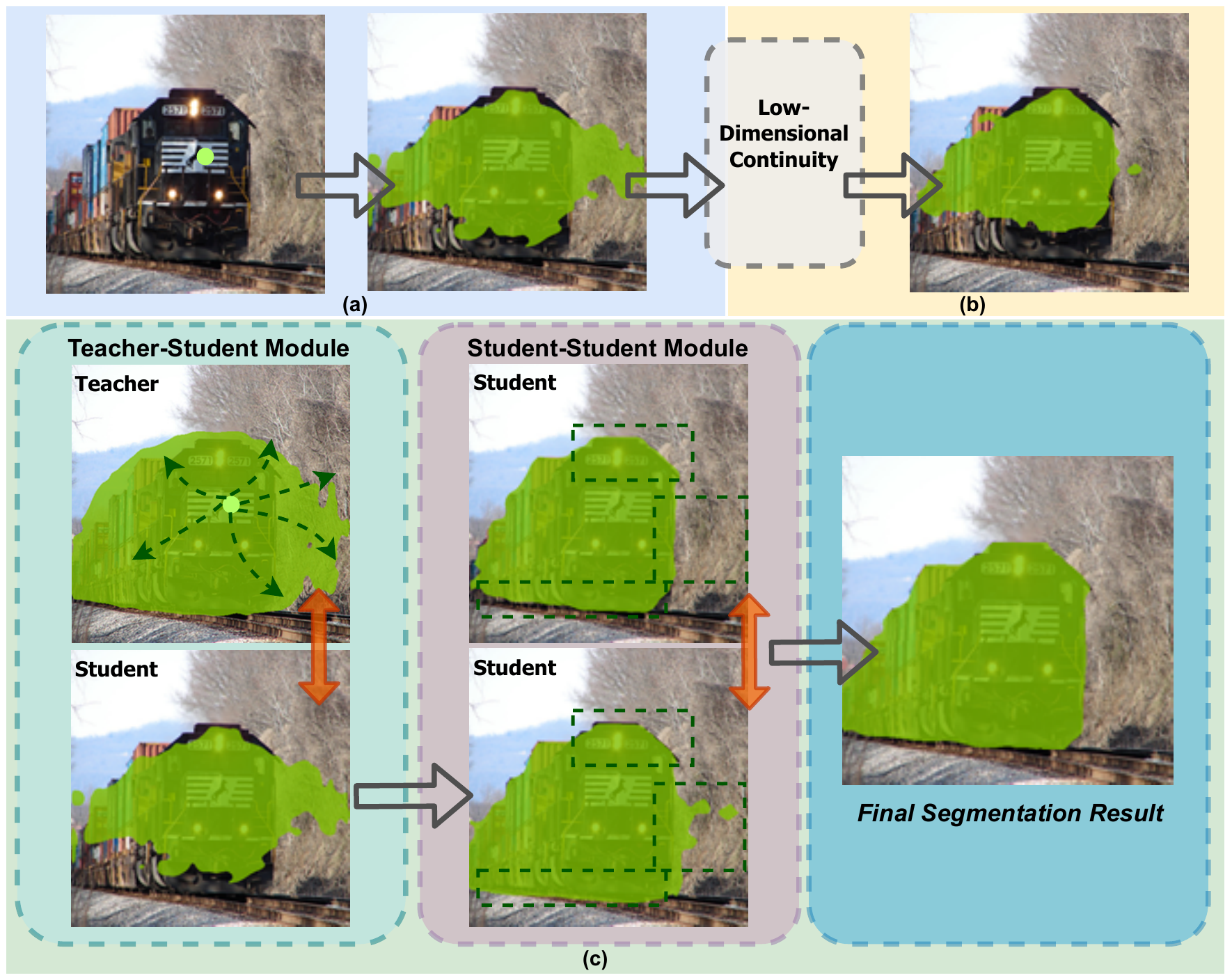}
	\end{center}
	\caption{Weakly supervised segmentation with the click-level annotations. (a) A generic model trained only with click-level annotations overfits to the labels and cannot recognize the whole object. (b) Previous works (e.g. regularized loss) apply low-dimensional continuity information to the training, also failing to correctly segment the object. (c) With seminar learning, our teacher-student module enables the network to generalize to the whole object, as indicated by the arrows. Meanwhile, by integrating diverse information from the two networks, the student-student module can smooth the boundary area in the marked boxes.}
	\label{fig:head}
\end{figure}

In order to alleviate the burden of annotations, weakly supervised semantic segmentation has become increasingly popular, as it only requires coarse annotations, such as box-level \cite{dai2015boxsup}, image-level \cite{2015Weakly}, scribble-level \cite{2016ScribbleSup}, or click-level \cite{2016What} supervision. Among these, click-level supervision only annotates one pixel for each object in an image. Further, it not only provides valuable location information, but is also one of the cheapest form of weakly supervision \cite{2016What}. It has high research potential in terms of the trade-off between information and time costs.

As commonly known, it is challenging to achieve satisfactory performance with limited supervision information during model training.
For instance, with click-level patterns, if only learning from one labeled pixel, the model cannot infer the entire range of an object, especially the edges, which will eventually weaken the segmentation performance.
An effective way to compensate weak supervision information is to introduce more prior information. For example, `What's the point' \cite{2016What} incorporates an objectness prior into network training, which helps distinguish between foreground and background. `ScribbleSup' \cite{2016ScribbleSup} uses an additional graphical model to propagate information from click-level annotations. `Regularized Loss' \cite{tang2018regularized} designs a regularization item based on dense conditional random field (CRF) for classifying nearby pixels with similar colors into the same category. 
These models only focus on low-dimensional continuity between labeled pixels and others pixels, which is limited to local annotation information in click-level supervision.
Therefore, these models cannot properly segment the entire object and still underperform.

Considering the nature of click-level supervised semantic segmentation, we make two observations: 1) A large number of unlabeled pixels are not well used, but could provide broader information, which can expand the learning range of networks from a single annotated pixel to an entire object. 2) If a network is trained under different conditions, such as using different random seeds, the predictions will vary greatly. This uncertainty causes that different networks capture distinctive and diverse information, which could be aggregated to complement each other.

Inspired by these observations, we propose \textit{seminar learning}, a novel learning paradigm for click-level weakly supervised semantic segmentation by introducing more effective information. The essence of our seminar learning is to complement the deficiency of networks by leveraging the knowledge provided from the predictions of other networks. As shown in Fig. \ref{fig:head}, seminar learning framework consists of two components: teacher-student module and student-student module.
Notably, the teacher-student module is exploited to expand the learning range of networks. We use an exponential moving average (EMA) based teacher network for generalized prediction and prevent the student network from overfitting to click-level labels, which has a similar workflow to semi-supervised mean-teacher \cite{2017Mean} method. However, compared to mean-teacher, our module is able to operate on unlabeled pixels in each image instead of unlabeled images. The student-student module is applied to refine segmentation boundaries by aggregating diversity information of student networks. To improve the efficiency of information transfer, we propose heterogeneous pseudo-labels as bridges between student networks, which based on the prediction of a fully trained student to guide the other.
In summary, we make several major contributions as follows: 
\begin{itemize}
    \item We propose a novel learning paradigm, called seminar learning, that can learn to leverage more supervisory information provided by a group of networks.
    
    \item We treat the click-level supervised semantic segmentation task as a semi-supervised pixel classification task per image, and propose a novel pixel consistency loss, which enables a student to learn from a teacher using unlabeled pixels. 
    
    \item The novel concept of heterogeneous pseudo-labels is proposed, which is a more effective medium to enable the supervisory information to be shared among diverse networks by the student-student module. 
    
    \item We conduct extensive experiments to verify the effectiveness of the proposed seminar learning, which outperforms previous SOTA works \cite{tang2018regularized} by a large margin (from 55.63\% to 72.51\% in terms of the mIOU metric).
\end{itemize}

\begin{figure*}[ht]
\begin{center}
	\includegraphics[width=0.9\linewidth]{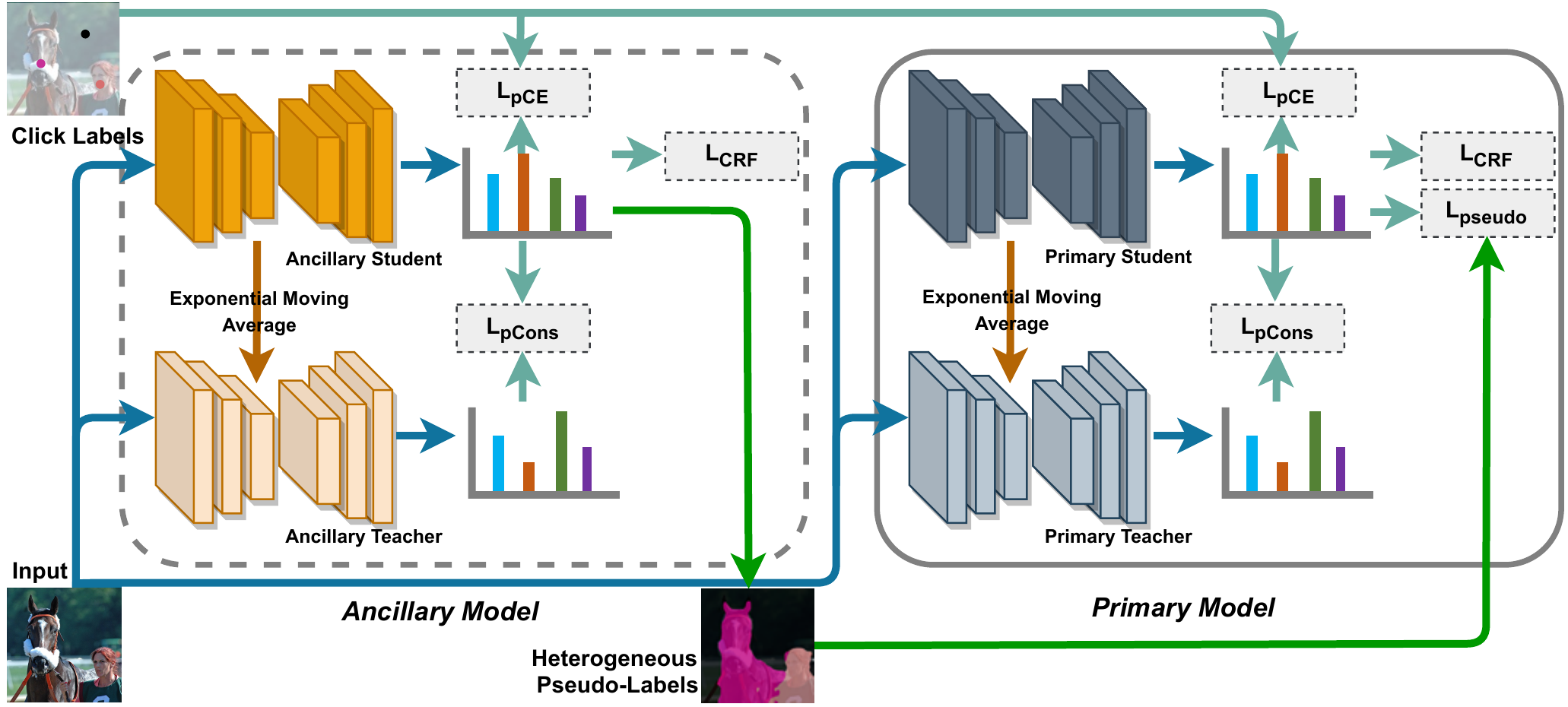}
\end{center}
\caption{The pipeline of the proposed seminar learning method for click-level supervised semantic segmentation. It consists of a primary model and ancillary model, which are trained progressively. 
}
\label{fig:pipeline}
\end{figure*}

\section{Related Work}

\textbf{Semantic segmentation.} Semantic segmentation can be viewed as pixel-wise classification, where each pixel is assigned to a category. Using deep learning in semantic segmentation began when the fully convolutional network (FCN) \cite{long2015fully} first arose. Most popular architectures are based on encoder-decoder models like FCN, such as SegNet \cite{badrinarayanan2017segnet}, U-Net \cite{ronneberger2015u}, MobileNet \cite{sandler2018mobilenetv2} and RefineNet \cite{lin2017refinenet}. Recent works have made great efforts to increase the receptive field of the network. For example, Chen \textit{et al.} \cite{chen2018encoder} extract long-range information without down-scaling the image by atrous convolutions. DeepLabv2 \cite{chen2017deeplab} introduces an atrous spatial pyramid pooling module.
In this paper, we employ DeepLab-v3+ \cite{chen2018encoder} as our backbone since it integrates various effective modules and outperforms previous models.

\textbf{Semi-supervised learning.} In general, semi-supervised learning tackles the training problem with only a small amount of labeled data. How to use plenty of unlabeled data becomes crucial for semi-supervised learning. A effective method is pseudo-labeling \cite{lee2013pseudo}, which use the prediction generated by itself to supervise unlabeled data. Recently, there are many semi-supervised methods based on the conception of predictions consistent with perturbations. 
The $\Pi$-model \cite{laine2016temporal} produces two noisy models, named the student model and the teacher model, and applies a consistency loss on the teacher and student predictions to adapt the model to the noise. Virtual adversarial training \cite{miyato2018virtual} is similar to the $\Pi$-model but it uses adversarial perturbation as noise. Mean-teacher \cite{2017Mean} improves the teacher-student architecture. It obtains the teacher model by moving the average of the student model weights. This practice has also been followed by subsequent semi-supervised works \cite{berthelot2019mixmatch, luo2018smooth, verma2019interpolation, sohn2020fixmatch}.
Besides, many algorithms utilize these semi-supervised methods for the image segmentation problem. For instance, \cite{choi2019self} directly applies mean-teacher to the unlabeled data in semi-supervised semantic segmentation, while \cite{luo2020semi} using the mean-teacher method between a strong label and a weak label.

\textbf{Weakly-supervised semantic segmentation.} Collecting accurate pixel-wise semantic labels is a labor-intensive process \cite{2016What}. To reduce the cost, weakly supervised annotations has been proposed, including click-level \cite{2016What, qian2019weakly}, scribble-level \cite{2016ScribbleSup, tang2018regularized, tang2018normalized}, box-level \cite{dai2015boxsup, khoreva2017simple}, and image-level supervision \cite{ahn2018learning}.
Although image-level labels require the least effort to collect, they contain limited information and are not suitable for complex scenarios. Recent works \cite{kim2021discriminative, chen2020weakly, lee2019ficklenet} usually use class activation maps (CAM) \cite{zhou2016learning} to generate labels for training.
Box-level annotations carry the most object information but also require the most effort to collect. Most existing methods utilize bounding boxes as search regions to conduct low-level searches for object masks \cite{dai2015boxsup, song2019box, kulharia2020box2seg}.
The collection costs of click-level annotation is between image-level and box-level, providing a trade-off in cost and information. However, few methods have tackled the problem of click-level segmentation algorithms since it is introduced in \cite{2016What}. Scribbles can be considered as an extension of click-level labels so that scribble-level supervised methods usually tests their performance under click-level supervision. The previous works \cite{2016ScribbleSup, 2016What} tried to mimic their fully-supervised counterparts by generating proposals. Specifically, they generate proposals by GrabCut \cite{rother2004grabcut} and alternate between the CNN training and proposal generation steps. This iteration method is tailored to their specific settings, which limits its generalization. To solve this problem, \cite{tang2018normalized} introduced regularized losses inspired by the normalized cuts method. Then \cite{tang2018regularized} improved the regularized losses with a dense CRF loss. These methods are effective under scribble-level supervision but still struggle with click-level supervision.

\section{Methodology}

In this section, we will provide a detailed description on our proposed seminar learning for click-level supervised semantic segmentation. Our framework mainly consists of the teacher-student and student-student module, which used to transfer information among networks. The combination of the two modules is similar to a real-world seminar, which was the inspiration of seminar learning. We will describe the overall process first and then explain how it works.

\subsection{Seminar Learning}

An overview of our proposed approach is shown in Fig. \ref{fig:pipeline}. We train the ancillary model first, and then the primary model. For each model, we apply a teacher-student module.

Meanwhile, heterogeneous pseudo-labels generated by an ancillary student are used as the extra input of the primary model, which constitute the student-student module. In this way, the primary model can integrate information from the ancillary model. 

A unified CNN framework is used for training. We define input pairs of images as $X$, of size $W \times H$, with corresponding annotation $\hat{Y}$; $x$ and $\hat{y}$ as the pixels of $X$ and $\hat{Y}$; $N = W \times H$ as the total number of pixels in each image; and $n$ as labeled pixels of each image in our click-level supervised task. The network outputs a softmax score map $Y$ of size $W \times H \times C$, where $C$ is the number of label classes. For the test process, the score map chooses the class of the max score for each pixel, and a final prediction of size $H \times W$ is obtained.

The training procedure can be described as follows:

\textbf{Training the ancillary model.} 
The ancillary model is constructed by the teacher-student module. In this module, we only need to train the student network. The teacher network is obtained by the exponential moving average (EMA) of the student network. At the training iteration $ t $, the EMA process is defined as
\begin{equation}\label{eq:moveaverage}
    \theta^{'}_t = \left\{
    \begin{array}{lcl}
    (1-\frac{1}{t}) \times \theta^{'}_{t-1} + \frac{1}{t} \times \theta_{t}, & & 1-\frac{1}{t} < \alpha \\
    \alpha \theta^{'}_{t-1} + (1 - \alpha)\theta_{t}, & & otherwise,
    \end{array}
    \right .
\end{equation}
where $ \alpha $ is a smoothing coefficient hyperparameter, and $ \theta^{'} $ and $ \theta $ are the weight of the teacher and student, respectively. To renew the weight of the teacher model quickly during the initial training iterations, we use absolute average instead of EMA when $1-\frac{1}{t} < \alpha$.

The networks of the ancillary student and the ancillary teacher are randomly initialized with the same random seed. In each iteration of the training, we input training images to both the student and teacher network and three losses are evaluated. Firstly, we train the student network using click-level labels by minimizing the partial cross-entropy loss $L_{pCE}$ \cite{tang2018regularized}, which is defined as 
\begin{equation}\label{eq:ce}
   L_{pCE} = - \frac{1}{n}\sum_{i \in n}\hat{y}^c_i\log(y^c_{i}),
\end{equation}
where $ i \in n $ indicates that only labeled pixels participate in the calculation of the loss, and $ \hat{y}^c_{i} = [0, 1]^c $ is the ground truth of pixel $ i $ belonging to class $ c $. 

To obtain the assistance of the ancillary teacher network, we apply a pixel consistency loss $L_{pCons}$, which is defined as 
\begin{equation}\label{eq:consistencyloss_pixel}
    L_{pCons} = - \frac{1}{N}\sum_{i \in N}||f(x_i, \theta^{'}) - f(x_i, \theta)||^2,
\end{equation}
where $ f(\cdot) $ is the softmax prediction of the network and no gradient is calculated in the teacher network. 

An regularized loss $L_{CRF}$ \cite{tang2018regularized} is also applied to smooth the segmentation, which is defined as 
\begin{equation}
    L_{CRF} = \sum_{C}Y^{C'}W_{pq}(1-Y^C),
\end{equation}
where $W_{pq}$ is a dense Gaussian kernel with a role of the relaxation of dense CRF \cite{krahenbuhl2013parameter}, $Y^{C}$ is the softmax output of each class, and $Y^{C'}$ is the transposed matrix of $Y^{C}$. 

After the backpropagation of all the losses, the ancillary teacher network will be updated by EMA. This process continues iteratively until the end of the training. Overall, the ancillary student network is trained by loss $ L^* $, which is defined as
\begin{equation} \label{equ:L^*}
    L^* = L_{pCE} + \lambda_{pCons}L_{pCons} + \lambda_{CRF}L_{CRF},
\end{equation}
where $ \lambda $ controls the contribution of each loss term.

\textbf{Training the primary model.} 
In the primary model, we also use the teacher-student module for the training. In addition, we apply the student-student module to connect the ancillary student network and the primary student network with heterogeneous pseudo-labels. 

After the ancillary model is fully trained, the networks of the primary student and teacher are initialized in the same manner as their ancillary counterparts. During each training iteration, we train the primary student network the same way as the ancillary student network through EMA. Moreover, we input training images to the ancillary model, and obtain prediction maps. By choosing the maximal class of the prediction maps, we generate heterogeneous pseudo-labels to introduce the contribution of the ancillary student network to the training of the primary student network. To include the information of heterogeneous pseudo-labels, a new loss $L_{pseudo}$ is proposed. Since heterogeneous pseudo-labels are applied to each pixel, the loss is in the form of cross-entropy.  $L_{pseudo}$ is defined as
\begin{equation}\label{eq:pseudo}
    L_{pseudo}(\theta) = - \frac{1}{N}\sum_{i \in N}\tilde{y}^c_i\log(y^c_{i}),
\end{equation}
where $\tilde{y}$ denotes the heterogeneous pseudo-labels generated by ancillary model $\theta_{anc}$. 

Therefore, the primary student network is trained with the overview loss $L$ in each iteration, which is defined as
\begin{equation}
    L = L^* + \lambda_{pseudo} L_{pseudo}.
\end{equation}

\begin{figure*}[ht]
	\begin{center}
		\includegraphics[width=0.85\linewidth]{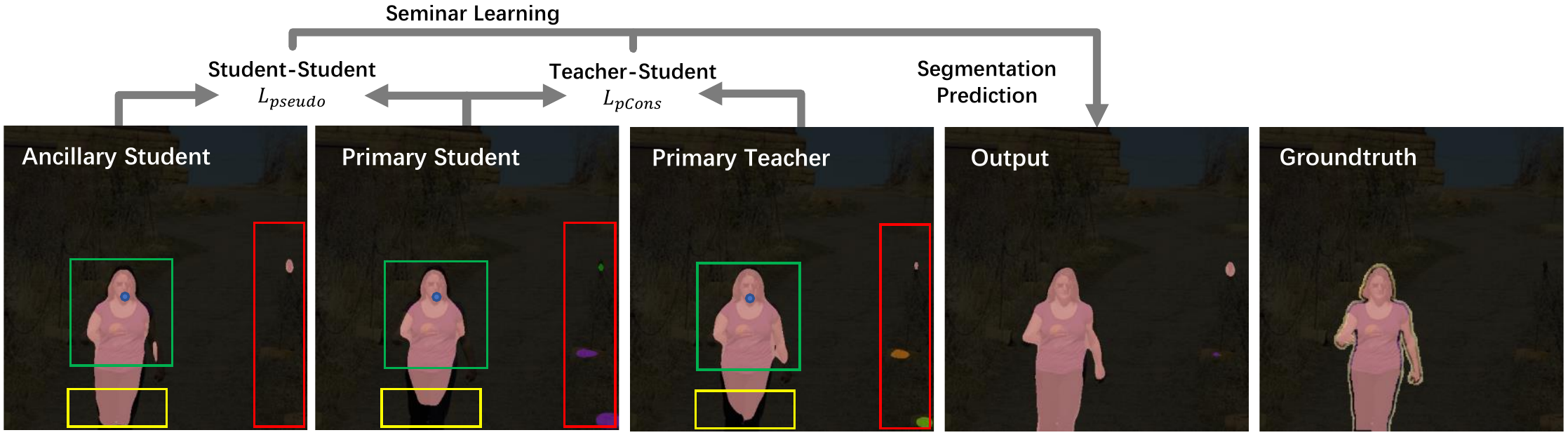}
	\end{center}
	\caption{Visualization of the mechanism in seminar learning. We obtain the first three results in the tenth epoch of primary model training.
	}
	\label{fig:seminardescription}
\end{figure*}

\subsection{Collaboration of modules}
\textbf{Teacher-student.} 
A large number of unlabeled pixels are not well utilized in click-level supervised semantic segmentation, which is also the case in semi-supervised learning (SSL). Thus, we regard the click-level supervision as a SSL task, where some image pixels are labeled while the others are unlabeled. The mean-teacher \cite{2017Mean} is a effective SSL method that use a teacher-student module to leverage unlabeled images. Inspired by this, we adapt the teacher-student module to our model by operating on unlabeled pixels instead of unlabeled images. 

In the teacher-student module, the teacher network is obtained by the EMA of the student network. The EMA network is proved to be more efficient than using the final network directly \cite{polyak1992acceleration}. EMA can be considered a temporal ensemble process, which endows it with a strong generalization ability. Thus, teacher network can avoid the overfitting to click-level labels and further guides the student network to learn the full object. In addition, its ability to reduce the bias of the targets can achieve a smoother classification boundary \cite{2017Mean}. Since the object boundary can be viewed as classification boundary \cite{french2020semi}, the EMA network can also predict a smoother and more accurate mask. 

To make consistency constraint between teacher and student network, we propose a pixel consistency loss $ L_{pCons} $, as a form of mean square error (MSE). Our pixel consistency loss only measures unlabeled pixels and is defined as:
\begin{equation}\label{eq:consistencyloss_pixel_former}
    \begin{aligned}
    L_{pCons} = \frac{1}{N-n}(\sum_{i \in n}||f(x_i, \theta^{'}) - f(x_i, \theta,)||^2 - \\  \sum_{i \in N}||f(x_i, \theta^{'}) - f(x_i, \theta)||^2).
    \end{aligned}
\end{equation}
Because $ n \ll N $, we ultimately use an approximate form of $ L_{pCons} $, defined in Eq. \ref{eq:consistencyloss_pixel}

\textbf{Student-student.} 
The teacher-student module is used as an individual model, while the student-student module is used to connect two models. In the student-student module, we propose heterogeneous pseudo-labels as the bridge between the two models. The heterogeneous pseudo-labels are generated by the ancillary student network after the network is fully trained. Then, the labels are transferred to the primary student network. 

Early attempts with pseudo-labels \cite{lee2013pseudo} used the network's predictions to train the network itself. However, such an operation will produce confirmation bias \cite{arazo2020pseudo}. In this case, the model will memorize the false pseudo-labels and it will be difficult to forget them during training. Therefore, we use a fully trained ancillary model to generate heterogeneous pseudo-labels. Such a discriminative model can produce reliable predictions that guide the training of the primary student network correctly.

Furthermore, the ancillary model should be trained under different conditions from the primary model, such as different random seed. As is mentioned before, models can generate different masks with great diversity. Learn from the prediction of the ancillary student network can compensate the deficiency of the primary student network and then smooth segmentation boundary. 

\subsection{Mechanism of Seminar Learning} 
We visualize the prediction of each network in the training of click-level supervised semantic segmentation, as shown in Fig. \ref{fig:seminardescription}, to illustrate why seminar learning works by leveraging teacher-student and student-student modules.

Comparing the segmentation of the ancillary student and primary student, we can see that the two networks have diverse predictions of the target person. Although the ancillary student fails to predict the right arm of the person in the green box, it has better robustness to noise in the red box and correctly predicts the legs of the person in the yellow box. The limbs of the person and background noise are uncertain regions since they are far from the click-level labels. By integrating the two networks in the student-student module, the ancillary student obtain a better segmentation performance in the leg and noisy regions. 

As for the teacher-student module, we find that the prediction of the primary teacher covers a wider region of the person in the green and yellow boxes compared to the primary student, which confirms that the primary teacher has better generalization. Since the primary teacher is updated by the primary student, the learning range of the primary student will gradually grow during training, finally allowing it to recognize the whole person.

After the training is done, we obtain the final segmentation prediction as the output. We can see that almost every part of the person is accurately predicted, and the final result is close to the ground truth. This shows that our seminar learning can effectively integrate information from all the networks in our pipeline and overcome the limitations of click-level labels to provide smoother segmentation.

\begin{table*}
\begin{center}

\begin{tabular}{l|c|c|l|c}
\hline
\tabincell{c}{Method} & \tabincell{c}{Foreground\\Annotation} & \tabincell{c}{Background\\Annotation} & \tabincell{c}{Specifics} & \tabincell{c}{mIOU (\%)}  \\
\hline\hline
What's the Point \cite{2016What} & manual & - & VGG16, size=[1$\times$1]px  & 43.40 \\
ScribbleSub  \cite{2016ScribbleSup} & synthetic & synthetic & Deeplab-v2-VGG16, size=[3$\times$3]px & 51.60 \\
Regularized Loss \cite{tang2018regularized} & synthetic & synthetic & Deeplab-v2-ResNet101, size=[3$\times$3]px & 57.00 \\
Regularized Loss \cite{tang2018regularized} & manual & synthetic & Deeplab-v3+-ResNet101, size=[1$\times$1]px & 55.63\\
\textbf{Ours} & manual & synthetic & Deeplab-v3+-ResNet101, size=[1$\times$1]px & \textbf{72.51} \\
\hline
\end{tabular}
\end{center}
\caption{Performance comparison on the Pascal VOC 2012 validation set. For all methods, we report performance under click-level supervision. We also describe the training specifics and the annotation source in the table.}
\label{table: Comparison of SOTA}
\end{table*}

\section{Experiments}
\subsection{Experimental Setup} \label{sec:Experiment setup}
\textbf{Implementation details.} We choose the network DeepLab-v3+ \cite{chen2018encoder} with ResNet101 \cite{he2016deep} as our backbone due to its flexible structure and excellent performance. In each model, we train the student network with a batch-size of 12 over 60 epochs. We follow the optimization strategy in DeepLab-V3+ for the baseline with full supervision, using stochastic gradient descent (SGD) \cite{bottou2012stochastic} with a base learning rate of 0.007, polynomial schedule \cite{liu2015parsenet}, momentum of 0.9, and weight decay of 5$e^{-4}$ for 90K steps. We set the hyperparameters of the loss weights as $\lambda_{CRF}=1$, $\lambda_{pseudo}=1$ and $\lambda_{pCons}=200$. These hyperparameters remain the same across all models. In each model, since the teacher network does not perform well in the early epochs, we ramp up the weight $\lambda_{pCons}$ from 0 to its final value during the first 40 epochs. The $\lambda_{pCons}$ is updated in each epoch with a linear function. We set the EMA decay parameter $\alpha=0.999$ for each teacher-student module. Our implementation is based on Pytorch \cite{paszke2017automatic}. All experiments are run on an Nvidia Titan RTX (24G) GPU.

\textbf{Datasets and annotations.} The majority of our experiments are on the PASCAL VOC 2012 segmentation dataset \cite{Everingham2010The}, which contains 10,582 training images, 1,449 validation images and 1,456 test images. We use the same setting as previous click-level supervision works. The mean intersections over union (mIOU) averaged over 21 classes are evaluated on the validation set. In previous works, there are two methods to obtain click-level labels for the Pascal VOC 2012 dataset. `What's the point' \cite{2016What} first proposed the concept of click-level supervised semantic segmentation, and manually annotated each object with only one pixel in an image. Each click-level label in \cite{2016ScribbleSup, tang2018regularized} is one point randomly selected from the scribble label of each object. We choose manual annotation as our training labels because they can better reflect the real click-level supervised annotations. However, the non-instance class (background) is not annotated in this annotation set. Thus we use the background labels proposed in \cite{obukhov2019gated}, which are synthetically generated from scribble labels. 

\textbf{Data augmentation.} Following \cite{tang2018regularized}, we only use the default data augmentation in DeepLab-V3+ to handle image data. It includes random scaling crop, horizontal flipping and random Gaussian noise for the training set and fixed scaling crop for the validation set. We set the image crop size to $513\times513$ for both the training and validation sets.

\subsection{Main Results}
Table \ref{table: Comparison of SOTA} compares our method against the SOTA weakly supervised approaches on the Pascal VOC 2012 validation set. In the table, the two `annotation' columns represent the sources of the click-level labels. `Manual' indicates that click-level labels are generated by manual annotation, and `synthetic' indicates that they are synthetically generated from scribbled labels. As mentioned in Sec. \ref{sec:Experiment setup}, our annotation set consists of manually annotated foreground labels and background labels generated from the scribble. The first three rows of the table are results published in the corresponding papers. In order to compare with the current SOTA method under the same conditions, we run the `Regularized Loss' method \cite{tang2018regularized} on the same annotation set, backbone and label size as our method. Its result is shown in the fourth row, which achieves 55.63\% mIOU when using the hyperparameter settings in its paper.  

Our method is based on the implementation of `Regularized Loss', which includes $L_{pCE}$ and $L_{CRF}$. Without modifying any parameters of the implementation in the fourth row, we apply our seminar learning method directly to `Regularized Loss'. After this operation, we greatly improve the mIOU from 55.63\% to 72.51\% (increased by 16.88\%), which far exceeds all previous SOTA methods.

\begin{table}[ht]
    \centering
    \begin{tabular}{c|c|c|c}
    \hline
        \tabincell{c}{Supervision \\ Level} & \tabincell{c}{Annotation Time \\ (sec/img)} & Method  & \tabincell{c}{mIOU \\ (\%)} \\
    \hline\hline
        Box     & 38.1 & BCM \cite{Song_2019_CVPR}       &70.2 \\
        Scribble& 34.9 & BPG \cite{wang2019boundary}     &73.2 \\
        Image  & 20.0 & DRS \cite{kim2021discriminative}&71.2 \\
        Click   & 22.1 & Ours                            &72.5 \\
    \hline
    \end{tabular}
    \caption{Performance comparison on the Pascal VOC 2012 validation dataset with various SOTA weakly supervised methods.}
    \label{tab:weaklycompare}
\end{table}

\textbf{Comparison with other weakly supervision.} In Table \ref{tab:weaklycompare}, we compare various levels of weakly supervised methods, all of which use ResNet101 as the backbone and no post-processing. The annotation times in the table are provided in \cite{2016What,bellver2019budget}. It can be seen that click-level supervision only acquires 22.1 seconds to annotate an image, which is close to image-level supervision. Meanwhile, our method gain a medial performance, which exceeds the image-level but is lower than the scribble-level, indicating that we make full use of the potential of click-level supervision and our method is a trade-off between time cost and performance.

\subsection{Ablation Study} 
\label{sec:ablatonstudy}

\begin{table}
    \setlength{\tabcolsep}{1mm}{
    \begin{center}
        \begin{tabular}{cccc|cc}
            \hline \small
            \multirowcell{2}{$ L_{pCE} $} & \multirowcell{2}{$ L_{CRF} $} & \multirowcell{2}{$ L_{pCons} $} & \multirowcell{2}{$ L_{pseudo} $} & \multicolumn{2}{c}{mIOU (\%)}   \\
            &&&&MobileNet&ResNet101\\
            \hline
            \hline
            
            \checkmark  &             &               &                & 50.85 & 54.70 \\
            \checkmark  & \checkmark  &               &                & 48.75 & 55.63 \\
            \checkmark  &             & \checkmark    &                & 56.14 & 61.47 \\
            \checkmark  & \checkmark  &  \checkmark   &                & 62.03 & 70.29 \\
            \checkmark  & \checkmark  & \checkmark    &  \checkmark    & 64.44 & 72.51 \\
            \hline 

             \multicolumn{4}{l}{Full supervision}                      & 71.92 & 78.59 \\
            \hline
        \end{tabular}
        \end{center}
    \caption{Performance comparison of our framework on PASCAL VOC 2012 val set. The experiments aim to verify the effects of each module in seminar learning on different networks. For all modules we report performance both under click-level supervision}
    \label{table: Comparison of framework}}
\end{table}

In this section, we verify seminar learning by testing the effectiveness of losses, student-student module, and hyperparameters separately.

\textbf{The effect of the losses in seminar learning.} 
This part aims to show the contributions of different losses on seminar learning. Table \ref{table: Comparison of framework} shows the mIOU scores of five combinations of losses for comparison. Notably, teacher network has the similar performance with student network after training, and we use student network as the output. In the combination without $L_{pseudo}$, the output of the ancillary student is final prediction.

In fact, `regularized loss' \cite{tang2018regularized} is the combination of $L_{pCE}+L_{CRF}$. By comparing 1st and 2nd rows of Table \ref{table: Comparison of framework}, $L_{CRF}+L_{pCE}$ cannot improve the segmentation performance evidently under the click-level supervision.

By comparing 2nd-4th rows of Table \ref{table: Comparison of framework}, we can find that $L_{pCE}+L_{CRF}+L_{pCons}$ provides a significant improvement where the mIOU grows from 55.63\% to 70.29\%, increasing by 14.66\%. Its result also exceeds that of $L_{pCE}+L_{pCons}$ by 8.82\%, which indicates that $L_{CRF}$ is helpful for improving the performance by cooperating with $L_{pCons}$. The 5th row shows that adding $L_{pseudo}$ to the training can receive an extra 2.22\% improvement and reach up to 72.51\%. 

Therefore, all the losses in seminar learning contribute to the improvement of the mIOU in click-level weakly supervised segmentation. Although adding $L_{CRF}$ to $L_{pCE}$ is not effect, $L_{CRF}$ can play a complementary role where intertwined with the pixel consistency loss $L_{pCons}$. 

To verify the generality of the proposed losses, we also test the losses on Deeplab-V3+ with MobileNet. The results of MobileNet have the same trend as that of ResNet101. It can prove that our losses can be used in other networks.

\begin{figure}[t]
\begin{center}
    \includegraphics[width=0.9\linewidth, trim=0 0 0 0]{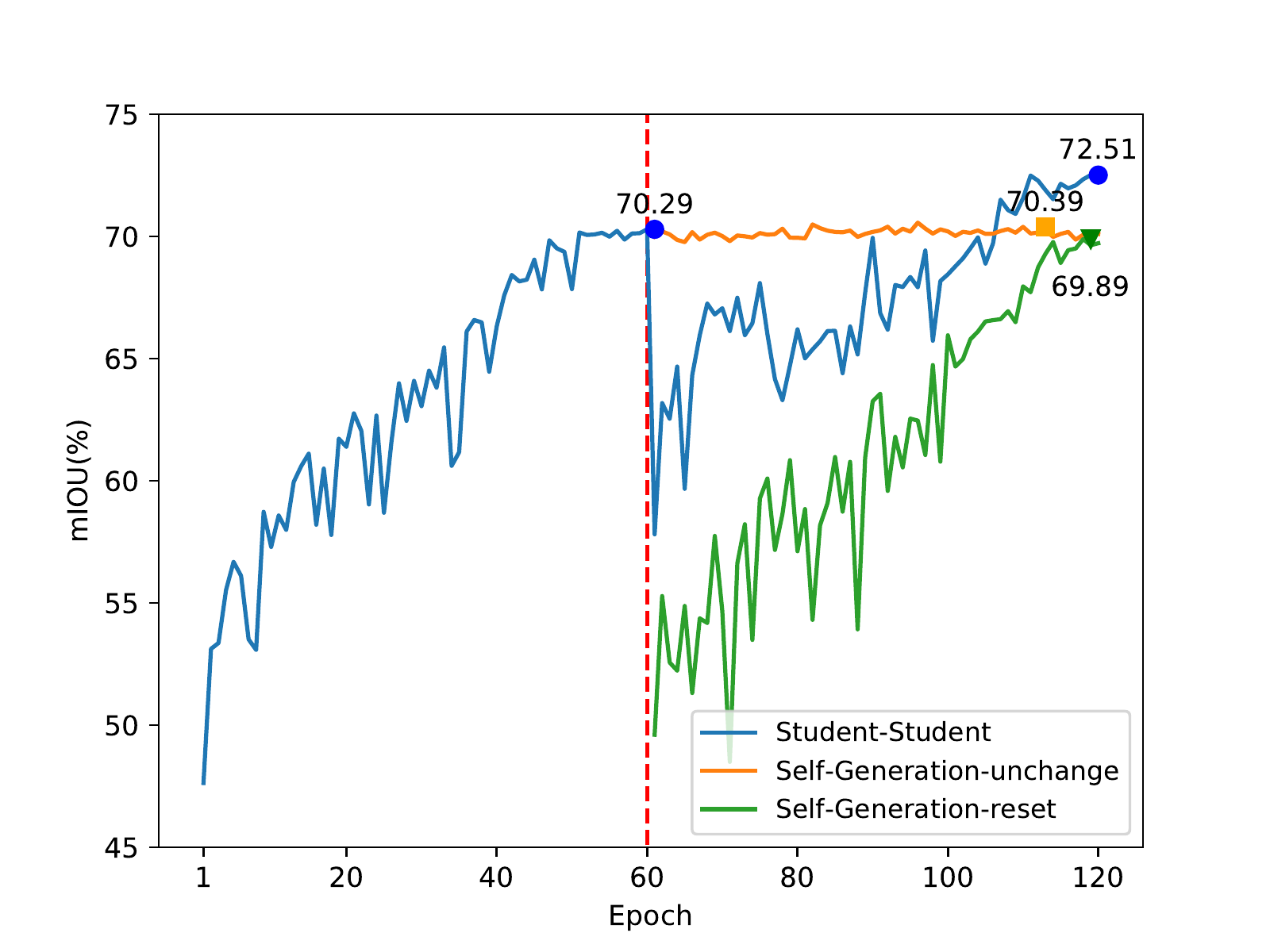}
\end{center}
  \caption{Comparison of different methods that use pseudo-labels on the Pascal VOC 2012 validation set. All results are obtained by DeeplabV3+ with ResNet101. }
\label{fig:proof}
\end{figure}

\textbf{The effectiveness of the student-student module.}
In student-student module, heterogeneous pseudo-labels are specially designed to transfer between two different networks. As shown in Fig. \ref{fig:proof}, we compare our module with two other cases to verify the effectiveness of student-student module. In the compared cases, the pseudo-labels are transferred to the same student network instead of a different student network, which we called self-generation method. Two cases use different learning rate schemes.

In the first 60 epochs, the three cases use the same ancillary model. We generate pseudo-labels in the 61th epoch. After the 60th epoch, we apply two learning rate schemes for the self-generation method: 1) Self-Generation-reset: we reset the learning rate at 61th epoch and keep the learning rate consistent with the student-student. It is obvious that the mIOU of the this case is always lower than that of the student-student module after training with pseudo-labels. The result becomes worse than that before the pseudo-labels are used. 2) Self-Generation-unchange: the learning rate remains unchanged after the 60th epoch. mIOU remains stable until the end of training. From the experiment, we find that the pseudo-labels generated by self-generation cannot improve the model. The student-student module is proved to be necessary to improve click-level supervised segmentation.

\begin{figure*}[htb]
\centering 
\subfigure[The number of student-student modules]{
    \label{fig:Pseudo_Iteration_LinearGraph}
    \includegraphics[width=0.28\textwidth]{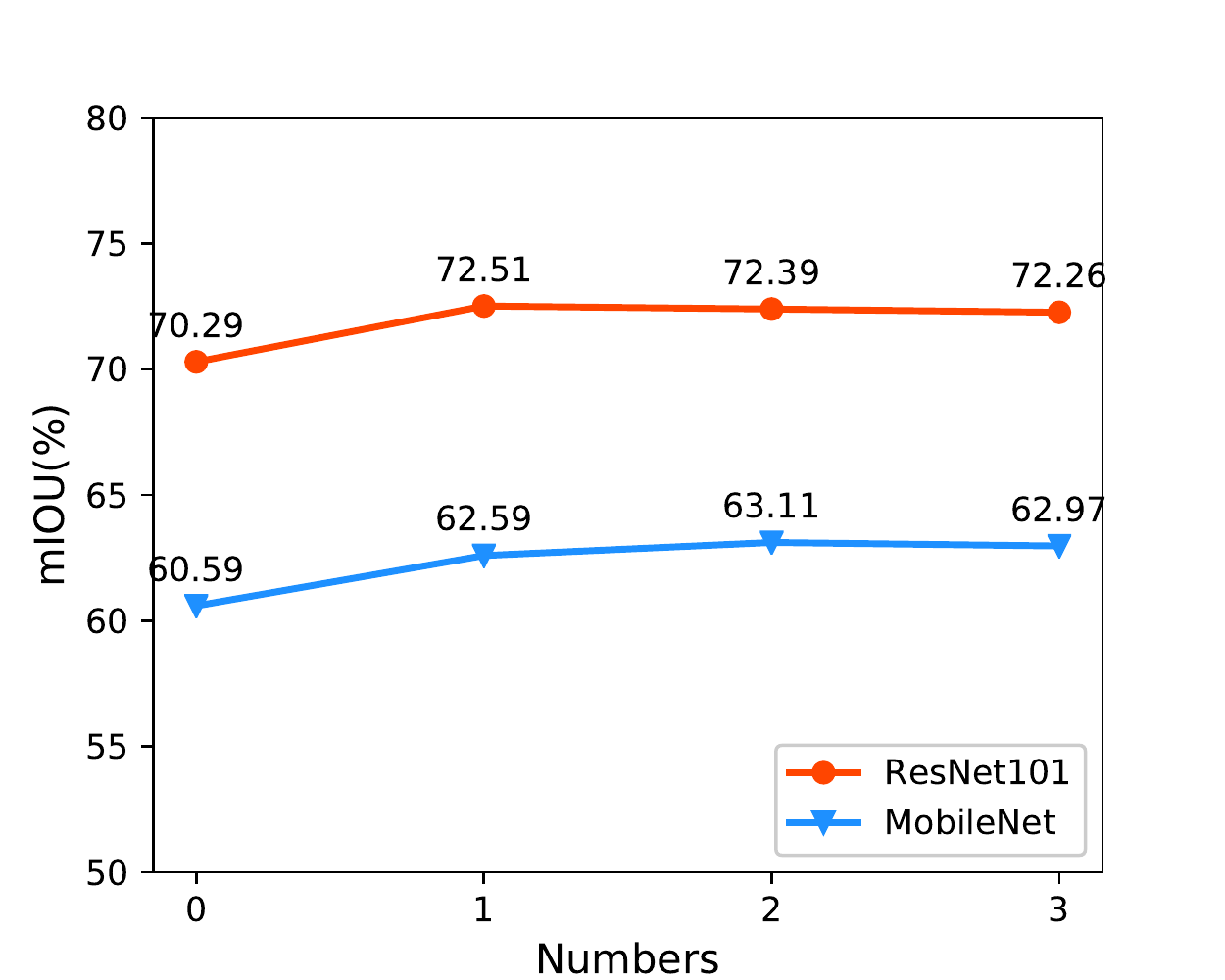}}
\subfigure[EMA decay weight]{
    \label{fig:EMA_Parameter_BarGraph}
    \includegraphics[width=0.28\textwidth]{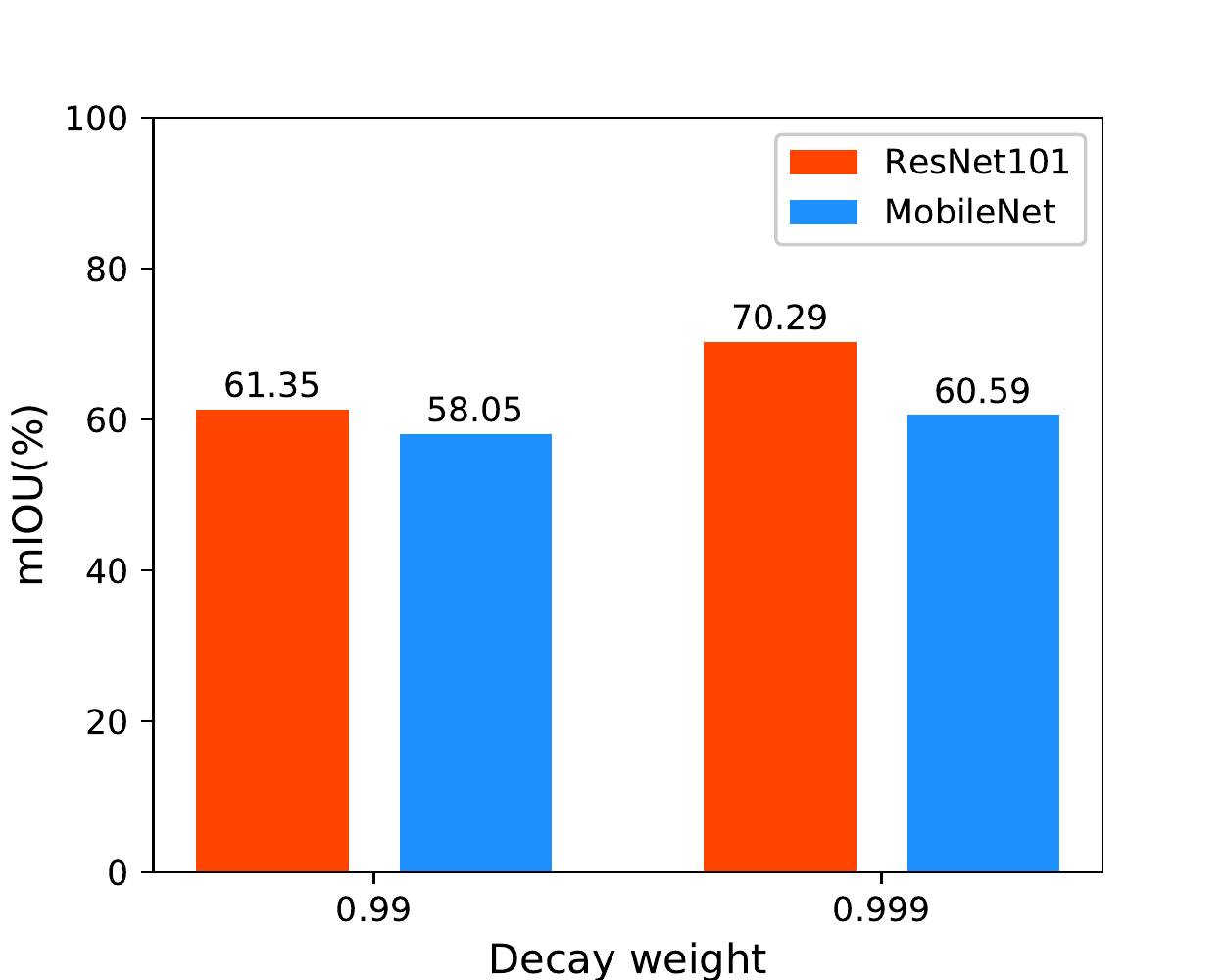}}
\subfigure[The weight of pixel consistency loss]{
    \label{fig:EMA_Consistency_Weight}
    \includegraphics[width=0.28\textwidth]{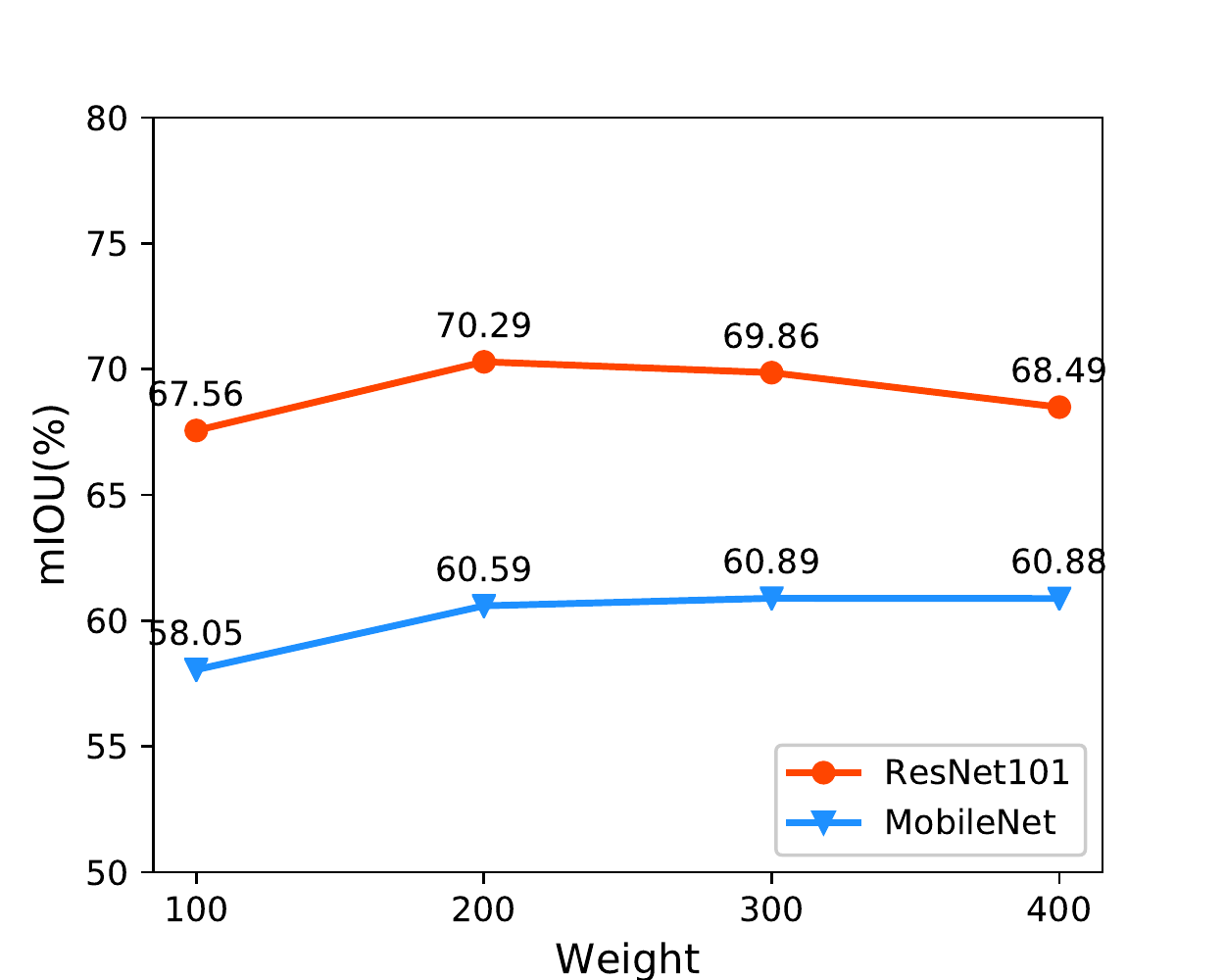}}
\caption{Ablation studies and empirical analysis on the Pascal VOC 2012 validation set. (a) Validation mIOU (\%) with different numbers of student-student modules; (b) Validation mIOU over the EMA decay weight $\alpha$; (c) Validation mIOU over weight $\lambda_{pCons}$ of the pixel consistency loss. }
\label{fig:train_step}
\end{figure*}

\textbf{The number of student-student modules.} 
The fully trained primary model can also act as the ancillary model for the next primary model. Therefore, our method can extend to a chain of student-student modules. Fig. \ref{fig:Pseudo_Iteration_LinearGraph} shows the performance in different numbers of student-student modules. We use two networks to evaluate how many student-student modules can improve the performance. When applying the first student-student module, the performance of the network is improved significantly. The accuracy of the Resnet101 improves from 70.29\% to 72.51\%, while that of the MobileNet improves from 60.59\% to 62.59\%. When we apply student-student modules more than once, the mIOU in both two networks almost does not increases. It concludes that one student-student module is sufficient for the training of networks.

\textbf{Sensitivity analysis of hyperparameters.} The performance of seminar learning depends on the hyperparameters of consistency weight $\lambda_{pCons}$ and EMA decay $ \alpha $. 

Fig. \ref{fig:EMA_Parameter_BarGraph} shows the sensitivity of decay weight $\alpha$ in the EMA. From \cite{2017Mean} we know that good decay always spans roughly an order of magnitude, and the commonly used hyperparameter values in current semi-supervised methods are 0.999 and 0.99. Our experiments prove that 0.999 is better in the click-level supervision task.

Fig. \ref{fig:EMA_Consistency_Weight} shows the sensitivity of the network to the consistency weight $\lambda_{pCons}$. We find that the value of $\lambda_{pCons}$ is much smaller than $\lambda_{pCE}$ of click-level supervision. This does not affect the training of the neural network. Thus, we increase the weight of the consistency loss $L_{pCons}$ to reach the same order of magnitude as $L_{pCE}$. Using this order of magnitude, we experiment with different hyperparameters on the two networks. We find that the student network can learn the best results from the teacher network when the consistency weight reaches 200.

\begin{figure}
\begin{center}
    \includegraphics[width=1\linewidth, trim=0 0 0 0]{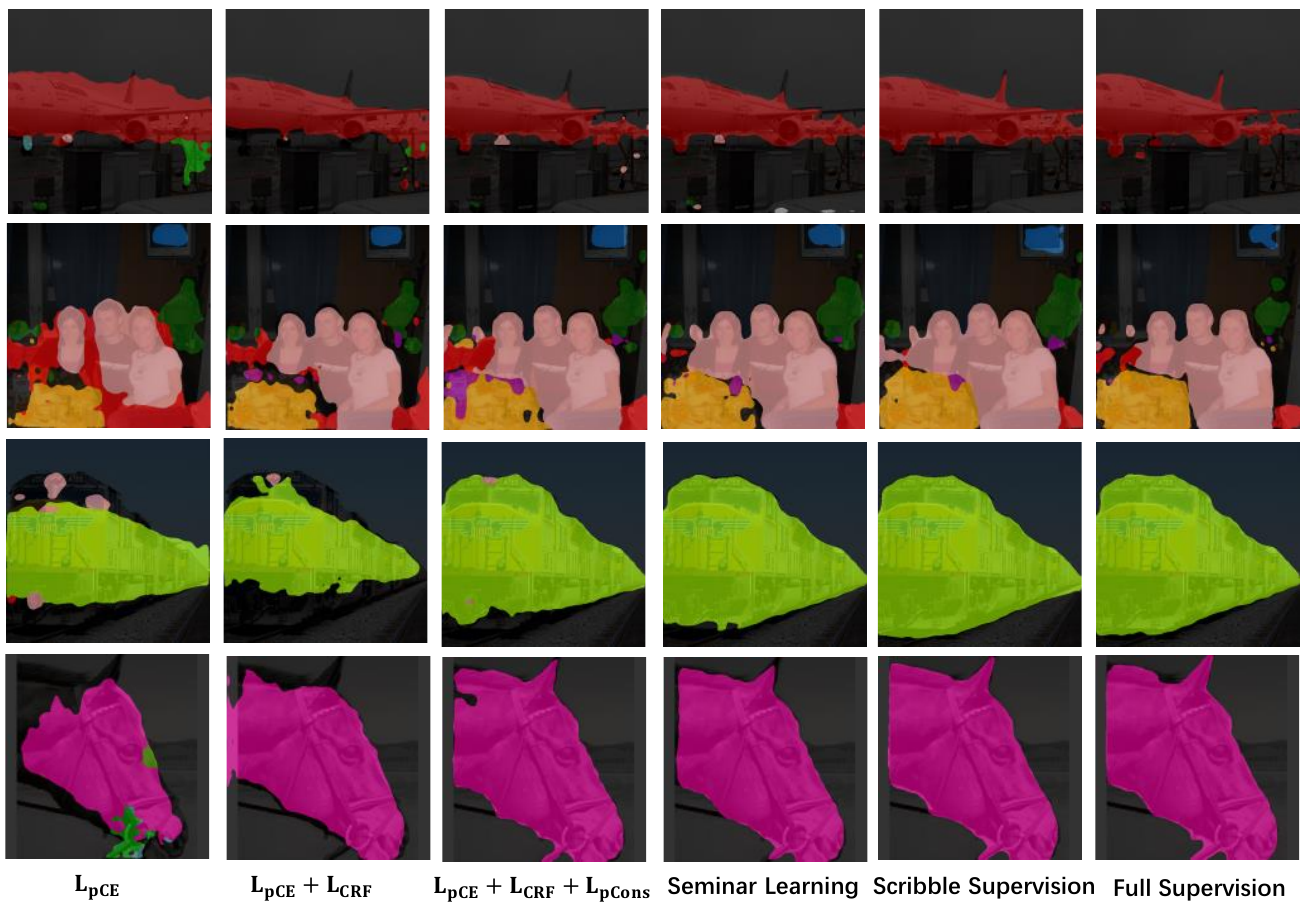}
\end{center}
  \caption{Visualization on the Pascal VOC 2012 validation set. }
\label{fig:Visualization}
\end{figure}

\subsection{Visualization}
Fig. \ref{fig:Visualization} shows examples of semantic segmentation predictions from qualitative views. In the first four columns, we apply different combinations of losses with seminar learning, and the segmentation results show the effects of them. 

\textbf{Visualized explanation of losses.}
As indicated in Sec. \ref{sec:ablatonstudy}, $L_{CRF}$ contributes to the performance only when it is combined with the pixel consistency loss $L_{pCons}$. 
Based on the theory of dense CRF \cite{krahenbuhl2011efficient}, if two pixels are close in both color and distance, the category association between them will be tight, and the two pixels are more likely to be predicted as the same category. 
Thus, the dense CRF based loss $L_{CRF}$ can make the segmentation result continuous inside objects. 
From the first two columns of Fig. \ref{fig:Visualization}, we can observe that the addition of $L_{CRF}$ enhances the internal continuity of the same category. 
However, as the limited labels contains insufficient information, the model fails to predict the pixels near the boundary of the object. 
Since $L_{CRF}$ is calculated with the prediction maps of the model, a large number of wrong predictions in the object prevent $L_{CRF}$ from establishing the correct connection among pixels, so $L_{CRF}$ does not work. 
In the third column of Fig. \ref{fig:Visualization}, after $L_{pCons}$ is applied, the boundary prediction is more accurate, and $L_{CRF}$ can smooth the predictions in the right way. 
Ultimately, our seminar learning achieve excellent performance with the combination of all losses.

\section{Conclusion}
In this paper, we propose a seminar learning paradigm for click-level weakly supervised semantic segmentation. Our approach consists of teacher-student and student-student modules, where we aggregate the generalized and diverse information from multiple networks. In this way, we address the insufficient information of the limited click-level labels. The experimental results demonstrate that our method achieves an excellent segmentation performance, surpassing the current SOTA model by 16.88\% (mIOU).

{\small
\bibliographystyle{ieee_fullname}
\bibliography{egbib}
}



\section*{Supplementary material}

# Supplementary Material: Seminar Learning for Click-Level Weakly Supervised Semantic Segmentation

Hongjun Chen[1], Jinbao Wang[1], Hong Cai Chen[1], Xiantong Zhen[2],
Feng Zheng[1*], Rongrong Ji[3], Ling Shao[4]
[1] Southern University of Science and Technology  [2] University of Amsterdam
[3] Xiamen University  [4] Inception Institute of Artificial Intelligence

## 1. Seminar learning in Scribble-Level Supervision

| Methods | mIOU(%) |
|---|---|
| Regularized Loss [2] | 75.18 |
| Ours | 76.17 |
| Full Supervision | 78.59 |

Table 1. Performance comparison on the Pascal VOC 2012 validation set. All results are evaluated on DeeplabV3+ and ResNet101.

Considering that scribble-level annotations can be treated as the extension of click-level annotations, we also apply the seminar learning to the scribble-level supervised semantic segmentation. We use the scribble annotation set first proposed in [1] to train our model. As shown in Table 1, we use 'regularized loss' method [2] as our baseline and gain 75.18% mIOU under our experimental settings. After applying the seminar learning, we gain a competitive result of 76.17% (0.99% higher). The results validate that our method has the potential to improve the performance of weak supervision at various levels.

Notably, since scribble-level labels are able to cross major regions of an object and carry stronger information, it is unnecessary to expand the learning range of the network by building a teacher-student network. However, uncertainties are still inevitable in such a setting as insufficient weak annotations covering the full objects. Therefore, we simplify our seminar learning pipeline for scribble-level supervision, excluding the teacher-student modules but retaining the two-stage training of the student-student module. specifically, the ancillary model is trained with the $L^*_{scribble}$ loss, which is defined as

$$L^*_{scribble} = L_{pCE} + \lambda_{CRF} L_{CRF}, \quad (1)$$

where $L_{CRF}$ is also an important regularization term for improving segmentation performance of scribble-level supervised semantic segmentation [2]. The overall $L_{scribble}$ loss for training the primary model is defined as

$$L_{scribble} = L^*_{scribble} + \lambda_{pseudo} L_{pseudo}. \quad (2)$$

## 2. Effect of Label Size

| Click Size | Only Use $L_{pCE}$ mIOU(%) | Seminar Learning mIOU(%) |
|---|---|---|
| $1 \times 1$ | 54.70 | 72.51 |
| $3 \times 3$ | 54.87 | 72.35 |
| $5 \times 5$ | 55.85 | 72.71 |

Table 2. The effect of enlarging size of the click-level labels. When the size is more than $1 \times 1$, click-level labels extend to their neighborhood.

As done by many existing approaches, we evaluate the sizes of click-level labels from $1 \times 1$ to $5 \times 5$ under our experimental setting. Table 2 conveys that increasing label size can slightly improve the segmentation performance. However, our seminar learning method gains the similar results in all sizes. It shows that our method is hardly affected by click-level label size.

## 3. More Visualizations

More examples of semantic segmentation in click-level weakly supervised semantic segmentation are shown in Fig. 1. We compare segmentation results among different combinations of losses, which shows the contribution of each loss function in our method.

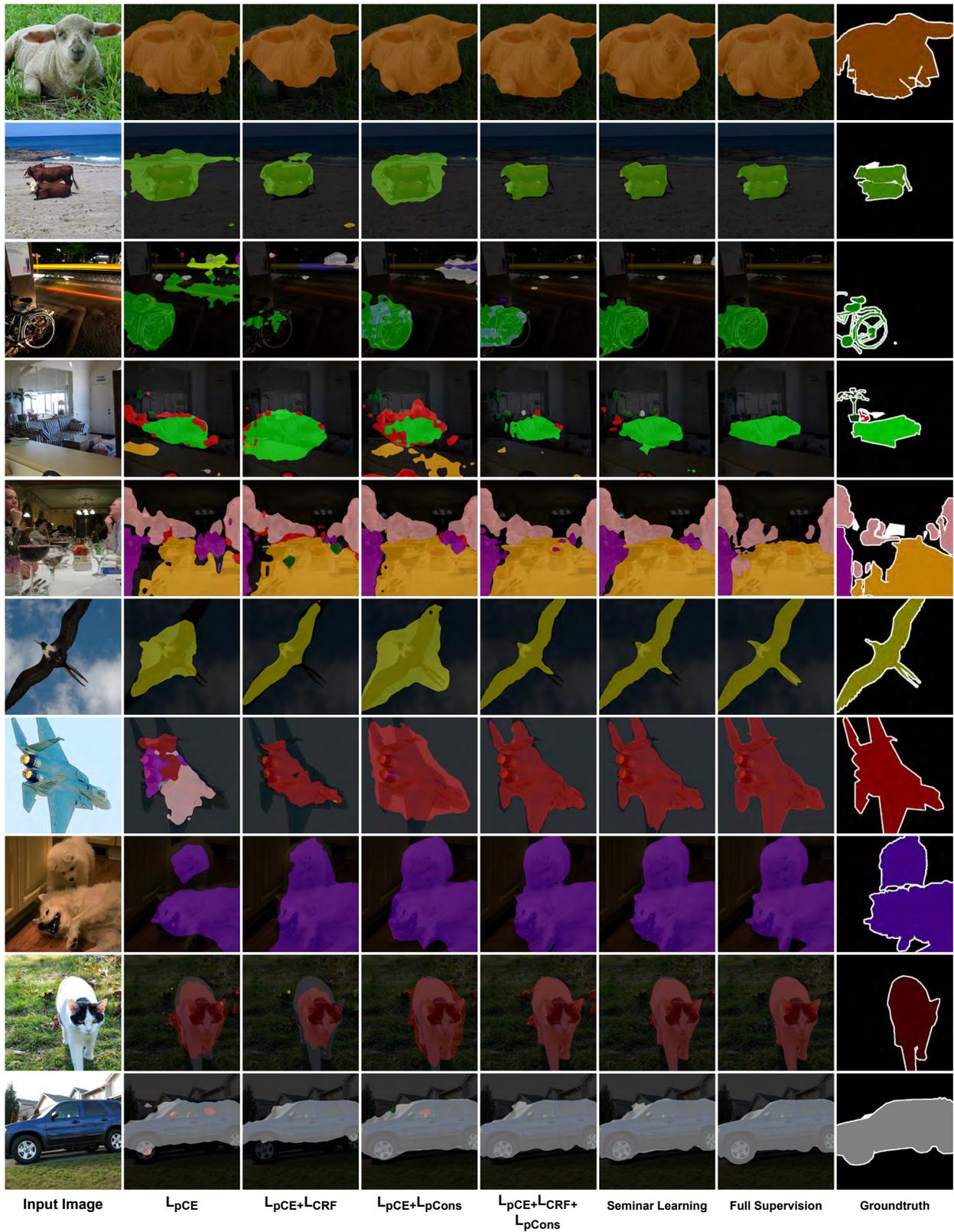

Figure 1. Qualitative results in various combinations of losses.



\end{document}